\documentclass[conference]{ieeeconf}
\IEEEoverridecommandlockouts

\usepackage{cite}
\usepackage{amsmath,amssymb,amsfonts}
\usepackage{algorithmic}
\usepackage{graphicx}
\usepackage{booktabs}
\usepackage{multirow}
\usepackage{adjustbox}
\usepackage{textcomp}
\usepackage{xcolor}
\usepackage{comment}
\usepackage{svg}
\usepackage{balance}
\usepackage{todonotes}
\include{bib_short.def}

\def\BibTeX{{\rm B\kern-.05em{\sc i\kern-.025em b}\kern-.08em
    T\kern-.1667em\lower.7ex\hbox{E}\kern-.125emX}}
\begin{document}
\newcommand{\TODO}[1]{{\textcolor{red}{{#1}}}} %
\title{\LARGE \bf BEVDriver: Leveraging BEV Maps in LLMs  \\ for Robust Closed-Loop Driving
}

\author{Katharina Winter$^{1}$, Mark Azer$^{1}$, Fabian B. Flohr$^{1}$
\thanks{$^{1}$Munich University of Applied Sciences, Intelligent Vehicles Lab (IVL), 80335 Munich, Germany
        {\tt\small intelligent-vehicles@hm.edu}}
}
\maketitle

\begin{abstract}
Autonomous driving has the potential to set the stage for more efficient future mobility, requiring the research domain to establish trust through safe, reliable and transparent driving. Large Language Models (LLMs) possess reasoning capabilities and natural language understanding, presenting the potential to serve as generalized decision-makers for ego-motion planning that can interact with humans and navigate environments designed for human drivers. While this research avenue is promising, current autonomous driving approaches are challenged by combining 3D spatial grounding and the reasoning and language capabilities of LLMs.
We introduce BEVDriver, an LLM-based model for end-to-end closed-loop driving in CARLA that utilizes latent BEV features as perception input. BEVDriver includes a BEV encoder to efficiently process multi-view images and 3D LiDAR point clouds. Within a common latent space, the BEV features are propagated through a Q-Former to align with natural language instructions and passed to the LLM that predicts and plans precise future trajectories while considering navigation instructions and critical scenarios. On the LangAuto benchmark, our model reaches up to 18.9\% higher performance on the Driving Score compared to SoTA methods.
\end{abstract}

\section{Introduction}

In the complex and safety-critical domain of autonomous driving, generative end-to-end methods are promising alternatives to rule-based or modular prediction and motion planning approaches, which exhibit shortcomings in their generalization capabilities. Furthermore, they typically scale well with more training data and optimize efficiently for the final driving task \cite{chen2024end}.
Due to their reasoning capabilities and natural language understanding, LLMs are considered proficient and interpretable decision makers. Their inherent world knowledge enhances context understanding and reasoning, while their ability to generate natural language explanations offers avenues for solving black box issues common for generative models. 
However, absolute spatial understanding has proven challenging for these models \cite{guo2024drivemllm, hwang2024emma}, representing a critical limitation in this safety-critical domain, where precise motion prediction and trajectory planning are crucial. 
Several transformer-based motion planning approaches \cite{shao2023reasonnet, yuan2024drama, weng2024paradrive, hu2022st} utilize Bird's-eye-view BEV feature maps to process sensor data from single or fused modalities into top-down feature representations depicting the surrounding environment of the ego-vehicle, providing a robust spatial input format to support tasks like trajectory planning \cite{li2023bev}.
With LMDrive, Shao et al. \cite{shao2024lmdrive} introduced an end-to-end driving approach capable of navigating the CARLA simulator \cite{dosovitskiy2017carla} using language instructions. Their approach leverages tokens for both perception -- such as object detection and traffic light status -- and future waypoint prediction, using a BEV encoder, highlighting the potential of BEV encoders as a promising direction for motion-planning research with LLMs. Building upon this foundation, our work advances the capabilities of LLMs by demonstrating that they can go beyond high-level decision making and refinement of predicted waypoints. 
We propose an LLM-based model, which predicts and plans trajectories end-to-end using BEV features from 3D LiDAR pointclouds and multi-view camera images while incorporating natural language instructions, effectively bridging high-level decision-making and low-level planning. 
While LMDrive \cite{shao2024lmdrive} relies on pre-predicted waypoint tokens as input, our approach eliminates this pre-conditioned, trajectory-specific processing by using raw latent BEV features of a perception-trained encoder-decoder vision model directly. 

This design fosters better generalization by allowing the LLM to interpret high-dimensional features without relying on pre-conditioned, trajectory-specific processing.

\begin{figure}[t]
    \centering
    \vspace*{3mm}
    \includegraphics[width=0.46\textwidth]{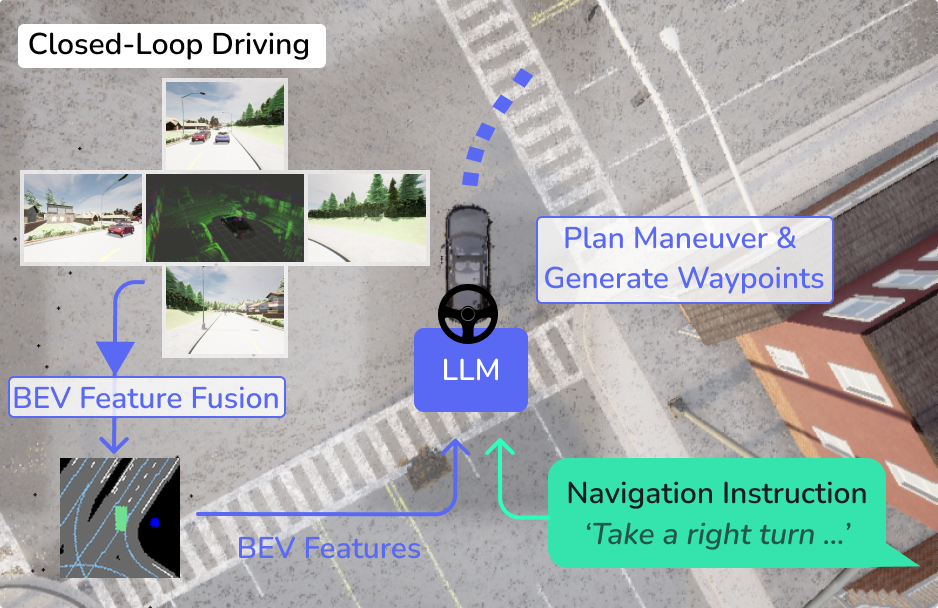} 
    \caption{We propose BEVDriver - an LLM-based end-to-end motion planner for closed-loop driving in CARLA. LiDAR and multi-view camera data are fused into BEV features based on which BEVDriver navigates the world following natural language navigation instructions combining high level planning and low level waypoint prediction.}
    \label{fig:eyecatcher}
\end{figure}

We summarize the key contributions of this work as follows. 
We introduce BEVDriver, an end-to-end LLM-driven model for closed-loop trajectory prediction and planning using raw BEV features from LiDAR and cameras (see Figure \ref{fig:eyecatcher}). Our approach removes reliance on pre-predicted waypoints, allowing the LLM to directly interpret high-dimensional BEV features for robust, adaptive planning. This improves generalization and enables self-supervised motion learning.
BEVDriver achieves SoTA performance on the LangAutoBenchmark, surpassing existing best methods by up to 18.9\% in closed-loop evaluations. We release model weights, training code, and dataset extensions on GitHub\footnote{https://github.com/intelligent-vehicles/BEVDriver} to support reproducibility and further research in language-guided autonomous driving.

\section{Related Work}

\paragraph{BEV Feature Maps}
BEV feature maps, derived from single or fused input modalities, offer a top-down view of the ego-vehicle's environment. 
This spatial representation enhances robustness against occlusions by providing a clear perspective of the surrounding environment \cite{li2023bev}. Moreover, this unified representation facilitates reasoning for downstream prediction and planning modules \cite{ng2020bev}. 
BEVDepth \cite{li2023bevdepth} focuses on 3D depth estimation for object detection using camera inputs. Multimodal BEV fusion can increase robustness of sensor failures, which is essential in this safety critical domain \cite{ng2020bev}. BEVFormer \cite{li2024bevformer} leverages Transformers with grid-based spatial queries for LiDAR-image fusion and temporal queries for sequential reasoning, while BEVFusion \cite{liu2023bevfusion} preserves semantic image information by integrating it into a shared BEV representation.

\paragraph{Closed-loop Driving} 
Closed-loop evaluation adjusts the ego-vehicle's actions based on environmental feedback, while open-loop evaluation measures deviation from a ground truth \cite{hallgarten2024can}.
Most state-of-the-art end-to-end methods focus on optimizing open-loop planning \cite{hu2023planning, hwang2024emma, sun2024sparsedrive, ding2024holistic, yang2024generalized}. However, open-loop evaluation is insufficient for assessing model performance, as success in open-loop does not ensure effectiveness in closed-loop scenarios \cite{sun2024sparsedrive, hwang2024emma, dauner2023parting}, emphasizing the need for dynamic, interactive evaluation.
The CARLA simulator offers a closed-loop evaluation environment, where BEV-based methods consistently rank among the top performers. InterFuser \cite{shao2023sinterfuser} enhances scene understanding by fusing LiDAR and multi-view camera inputs into a BEV projection, effectively detecting challenging scenarios. ReasonNet \cite{shao2023reasonnet} integrates temporal and global reasoning modules in BEV maps to capture dynamics and interactions.

\paragraph{LLM-based Driving}
LLMs are increasingly employed as motion planners in autonomous driving systems, utilizing world knowledge and reasoning abilities to generate context-aware and well-informed driving decisions, demonstrated by SoTA methods, like DriveLM \cite{sima2023drivelm}, DriveVLM \cite{tian2024drivevlm}, or EMMA \cite{hwang2024emma}.

Beyond improving reasoning and motion planning, natural language processing enhances interpretability—an issue in end-to-end generative models. LLM-based planners like DriveGPT4 \cite{xu2024drivegpt4}, RAG-Driver \cite{yuan2024rag}, and DriveMLM \cite{wang2023drivemlm} generate context-aware explanations learned from annotated datasets.

However, limited spatial understanding remains a key challenge and major drawback of LLM-based motion planning \cite{guo2024drivemllm, hwang2024emma}.
Existing methods like OmniDrive tackle this problem by using 3D-tokenization \cite{wang2024omnidrive}. CarLLaVA \cite{renz2024carllava} employs vision encoding to retain spatial features.

BEV maps can provide a robust, unified and spatial representation. Their efficiency and clarity make them an ideal solution, as shown in Talk2BEV \cite{choudhary2024talk2bev}, which integrates BEV maps with Vision Language Models (VLMs) to combine perception and reasoning for autonomous driving.

LLM-based methods have demonstrated effective closed-loop driving performance in CARLA. DriveMLM \cite{wang2023drivemlm} combines token embeddings from multi-view images, LiDAR pointclouds (processed by separate Q-Formers \cite{blip2}), system messages, and user instructions. The LLM generates high-level decisions (e.g., \textit{accelerate}, \textit{follow}, \textit{left\_change}) for the behavior planning module to produce control signals. 
AD-H \cite{zhang2024ad} is a hierarchical system of two LLMs, one representing a high level planner that produces driving commands for the other LLM functioning as low level controller, which produces future waypoints.
LMDrive \cite{shao2024lmdrive} uses BEV features from LiDAR and multi-view cameras to ground surrounding traffic and predict waypoints, which the LLM uses for decision-making based on natural language instructions.
While these methods use LLMs for high-level decisions or in a hybrid way, BEVDriver directly predicts waypoints using the LLM, without relying on a behavior planner.
Recently, CarLLaVA \cite{renz2024carllava} achieved impressive  results using only image inputs, though its single-modality approach may face robustness issues outside simulation. While the authors show that an LLM can generate adequate path waypoints for closed-loop driving, CarLLaVA omits natural language queries and thus lacks the broader contextual benefits that language can provide.
Our proposed model, BEVDriver, functions as a generalized agent by unifying accurate waypoint prediction from BEV features with natural language understanding.

\begin{figure*}[t]
    \centering
    \vspace*{2mm}
    \includegraphics[width=0.85\textwidth]{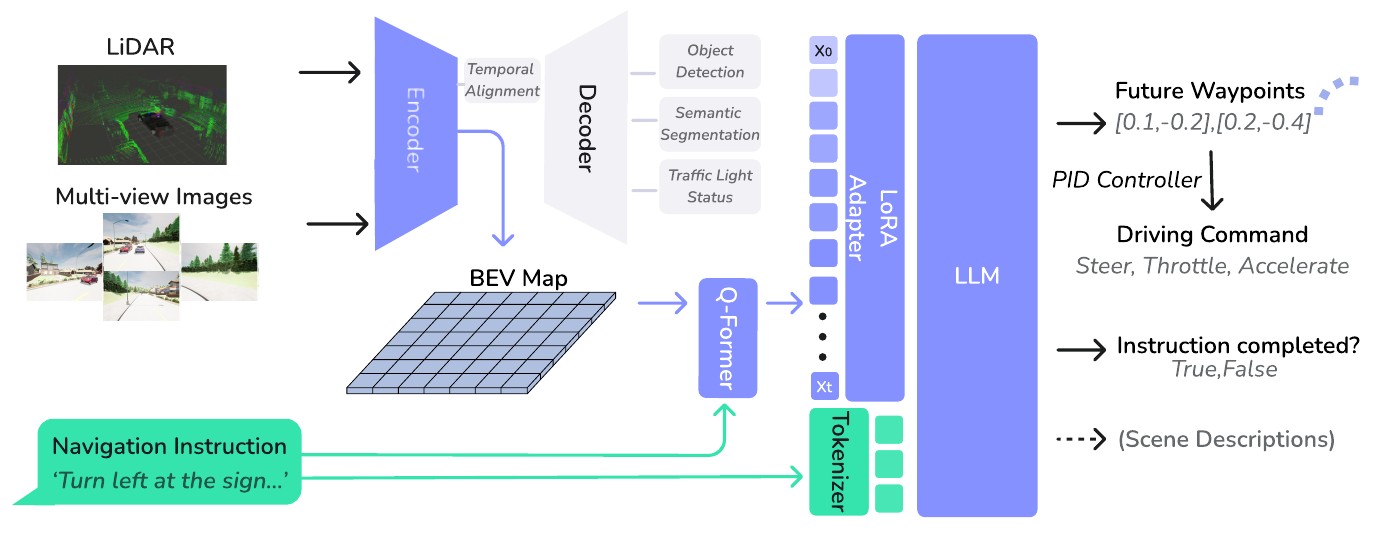} 
    \caption{Architecture of BEVDriver. Multi-view RGB images and 3D LiDAR point clouds are encoded into a BEV feature map, trained with object detection, semantic segmentation, traffic light detection and a self-supervised alignment loss. A Q-Former aligns the pre-trained latent features with the navigation instructions natural language space. A LoRA adapter feeds historical inputs to the LLM, which processes tokenized navigation instructions alongside perception data. The LLM outputs future waypoints, converted into driving commands by a PID controller, as well as scene descriptions and a boolean indicating instruction completion.}
    \label{fig:architecture}
\end{figure*}

\section{Method}

The overall architecture of our model is depicted in Figure~\ref{fig:architecture}. Four multi-view camera images and a LiDAR pointcloud are fused into BEV representation by the BEV encoder. This latent environment presentation is aligned with the navigation instructions in the natural language latent space by the Q-Former and projected into the LLM, which predicts and plans the future waypoints of the ego-vehicle using these latent inputs and the behavior-level navigation instructions. In the following, the architecture specification of our model is described in more detail.

\subsection{Perception}
For the generation of BEV feature embeddings we implement a BEV encoder adapting
InterFuser's \cite{shao2023sinterfuser} encoder architecture. This architecture utilizes ResNet-50 for image processing and ResNet-18 for LiDAR data, ensuring efficient multimodal feature extraction. The extracted features are then processed by a Transformer-based encoder that maps them to a 256-dimensional embedding space. On the decoder side, we apply InterFuser's traffic detection head, the traffic light status prediction head and add a front-view semantic segmentation head, providing a compact feature representation of the surrounding environment including grounding and semantic information.
The semantic segmentation head consists of a Pyramid Scene Parsing Network \cite{zhao2017pyramid} with adaptive average pooling at multiple scales (1, 2, 3, 6) to capture global context, followed by a projection layer that merges pooled features with the input features to preserve spatial details and convolutional layers for precise segmentation output. The BEV encoder is pre-trained with four losses: An L1 traffic detection loss, cross-entropy traffic light status, focal loss with weighted cross-entropy for segmentation ($\alpha=0.9$, $\gamma=5$), and contrastive loss for temporal alignment, promoting similarity between nearby frames while distinguishing distant ones.
During inference, we discard the decoder and propagate the latent BEV feature embeddings, yielding an efficient perception pipeline with an encoder size of 5.37M parameters and an inference frequence of 18Hz.

\subsection{End-to-End Pipeline}
The end-to-end driving pipeline consists of the trained BEV encoder which processes multiview camera images (front, left, right and rear view) and the LiDAR pointcloud to generate BEV latent embeddings. The BEV encoder processes historical sensor data for a maximum time horizon of 40 frames to capture temporal dependencies. The BEV embeddings are accumulated and refined by a modified BLIP-2 Q-Former \cite{blip2} adapted for the BEV inputs to be aligned with the navigation tokens in the natural language space employing 32 learnable query tokens with 768 dimensions. The LLM then processes the sequence of instruction and visual tokens through a LoRA adapter \cite{hu2022lora}, configured with rank 16 and a scaling factor of $32.5\%$ dropout, which efficiently adapts the interchangeable backbone LLM to generate action tokens that guide navigation. A GRU adapter network predicts five future waypoints using the last hidden layer of the LLM, while another 2-layer MLP adapter determines a completion flag indicating whether the given navigation instruction has been completed or an unfeasible or unsafe task is discarded. Finally, two PID controllers, one for longitudinal control and another for latitudinal control, regulate braking, throttling, and steering, ensuring precise tracking of the predicted waypoints and enabling accurate and instruction-compliant navigation.

\section{Experiments}

\subsection{Setup}
We employ the LLM Llama-7b \cite{touvron2023Llama} and the instruction-finetuned model Llama-3.1 8B-Instruct to compare BEVDriver in closed-loop evaluation using navigation instructions against LMDrive \cite{shao2024lmdrive} and AD-H \cite{zhang2024ad}, keeping the pre-trained BEV encoder frozen in the primary setup.
Input frames are sampled at a rate of 1 and the PID controller's integral error term is set to 20 to enhance control stability. Since the PID controller operates independently of the model, this setting can be applied universally across different architectures. All experiments are averaged over three repetitions per scenario.

\subsection{Datasets and Benchmark}
\paragraph{LMDrive Dataset}
We use the Dataset from LMDrive \cite{shao2024lmdrive} containing 15k sequences from routes of eight Carla Towns containing data with a sampling frequency of 10Hz including navigation instructions, control signals and front, left, right and rear view RGB images, LiDAR data and metadata about affordances, traffic light states and ground truth waypoint predictions. The training data contains all eight public CARLA towns (1,2,3,4,5,6,7,10), leaving the three weather-daytime conditions soft rain (noon), soft rain (sunset) and hard rain (night) for validation only.
\paragraph{BEVDriver Semantic Dataset}
We collect another 2k sequences on the eight towns used in LMDrive \cite{shao2024lmdrive}  that include ground truth semantic segmentation of the front, left and right view containing all CARLA semantic classes including pedestrians, vehicles, infrastructure (e.g., roads, sidewalks, traffic signs), and environmental features (e.g., buildings, nature). Our dataset and the code for data generation will be made publicly available.

\paragraph{CARLA Benchmark}
CARLA provides a closed-loop benchmark with its Leaderboard 1.0 challenge on various scenarios. 
The LangAuto Benchmark proposed by Shao et al. \cite{shao2024lmdrive} adapts the CARLA Leaderboard challenge for LLMs by mapping the seven discrete navigation commands to a variety of natural language navigation instructions, specifically designed for language model evaluation. 
On the eight public CARLA towns, LangAuto includes diverse environmental, weather and daylight conditions and challenging scenarios, such as intersections, sudden cyclist or pedestrian crossings, or multi-lane roundabouts. Just like the original CARLA Leaderboard, the benchmark is divided into different lengths of tiny ($<150$m), short ($150-500$m) and long routes ($>500$m) testing precision on basic driving maneuvers, reactive behavior and long-term consistency. In addition to the navigation instructions, misleading instructions containing unfeasible tasks are interspersed aiming to challenge the model's robustness. These misleading instructions are considered complete if the designated time passes without a response.
The evaluation includes unseen weather and daytime conditions.

\subsection{Metrics}
Closed-Loop Planning is evaluated based on three metrics, namely the Driving Score (DS), which is the product of Route Completion (RC), the percentage of the route completed, and the Infraction Score (IS), including collisions and other infractions like running a red light\footnote{ https://leaderboard.carla.org/}. 
We additionally provide an open-loop baseline utilizing the Average Displacement Error (ADE) and the Final Displacement Error (FDE) metrics, measuring L2 distance between prediction and ground truth averaged over all waypoints and at the final waypoint, respectively. The evaluation is conducted on the validation split of LMDrive using five future waypoints.

\subsection{Training}
The full pipeline is trained with a Llama-7b and a Llama-3.1 backbone. We trained both models with the AdamW optimizer with a weight decay of 0.06. A cosine learning rate scheduler was utilized, starting with an initial learning rate of 1e-4, a minimum learning rate of 1e-5, and a warmup learning rate of 1e-6 over 2000 warmup steps. Training was conducted for 15 epochs (72 hours) using eight Nvidia A100 GPUs with a batch size of 4.
 The BEV encoder is trained for 48 hours on eight Nvidia A40 GPUs with a batch size of 16, using AdamW with a learning-rate of 5e-4 and backbone learning rate of 2e-4 and weight decay of 0.05.
For the comparison with LMDrive on the LangAuto benchmark, we adopt their training and validation split.

\subsection{Qualitative Results}
\begin{figure}
    \centering
    \vspace*{2mm} 
    \hspace{-0.8cm}
    \includegraphics[width=0.4\textwidth]{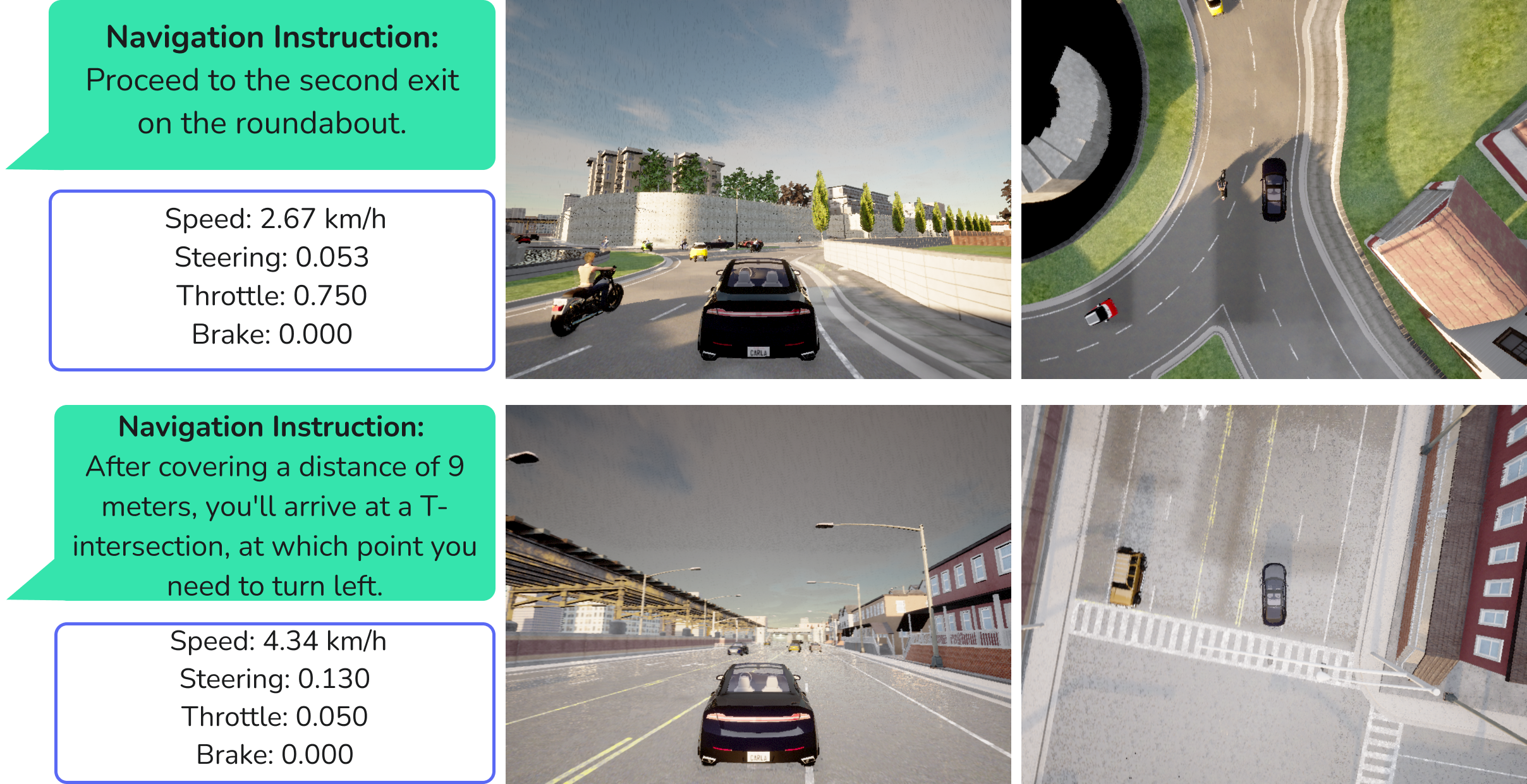} 
    \caption{Two qualitative samples of the model driving in the CARLA simulator in a third-person and top-down view following the navigation instruction.}
    \label{fig:qualitative}
\end{figure}

Figure \ref{fig:qualitative} displays two qualitative samples of the model navigating following the instructions. BEVDriver proficiently navigates complex street layouts like roundabouts. It also plans in foresight of an upcaming task, which is depicted by an early lane changing maneuver in the bottom sample.

\begin{figure}[t]
    \vspace*{2mm} 
  \centering
  \setlength{\tabcolsep}{2pt}
  \begin{tabular}{ccccc}
    \text{\small Front RGB} & 
    \text{\small Seg.\ GT} & 
    \text{\small Seg.\ Pred.} &
    \text{\small Det. GT} &
    \text{\small Det. Pred.} \\
    \includegraphics[width=0.08\textwidth]{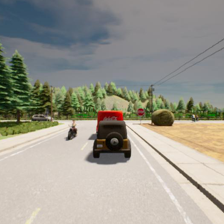} &
    \includegraphics[width=0.08\textwidth]{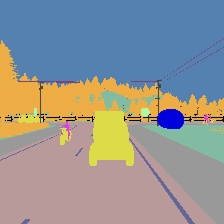} &
    \includegraphics[width=0.08\textwidth]{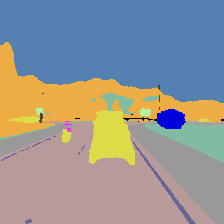} &
    \includegraphics[width=0.08\textwidth]{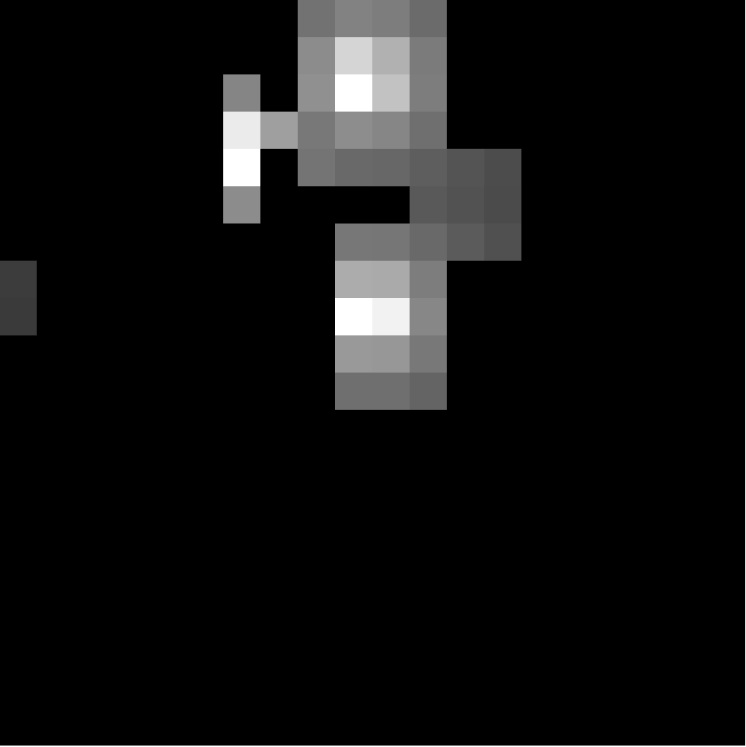} &
    \includegraphics[width=0.08\textwidth]{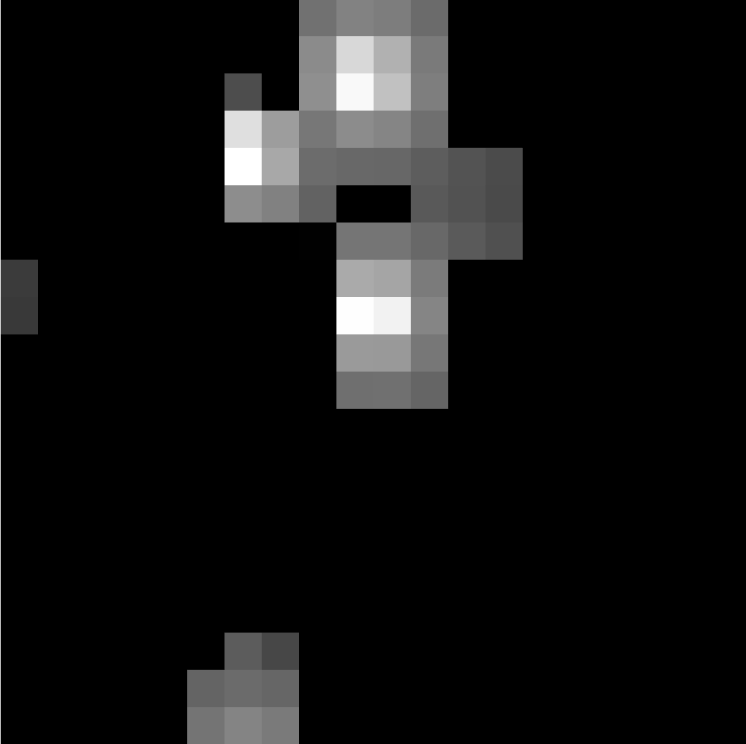} \\
    \includegraphics[width=0.08\textwidth]{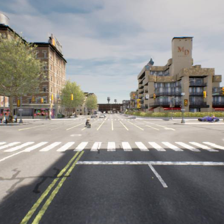} &
    \includegraphics[width=0.08\textwidth]{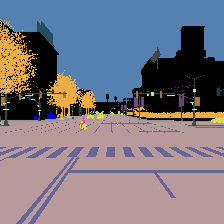} &
    \includegraphics[width=0.08\textwidth]{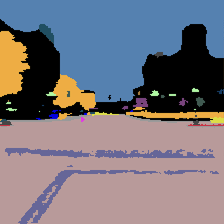} &
    \includegraphics[width=0.08\textwidth]{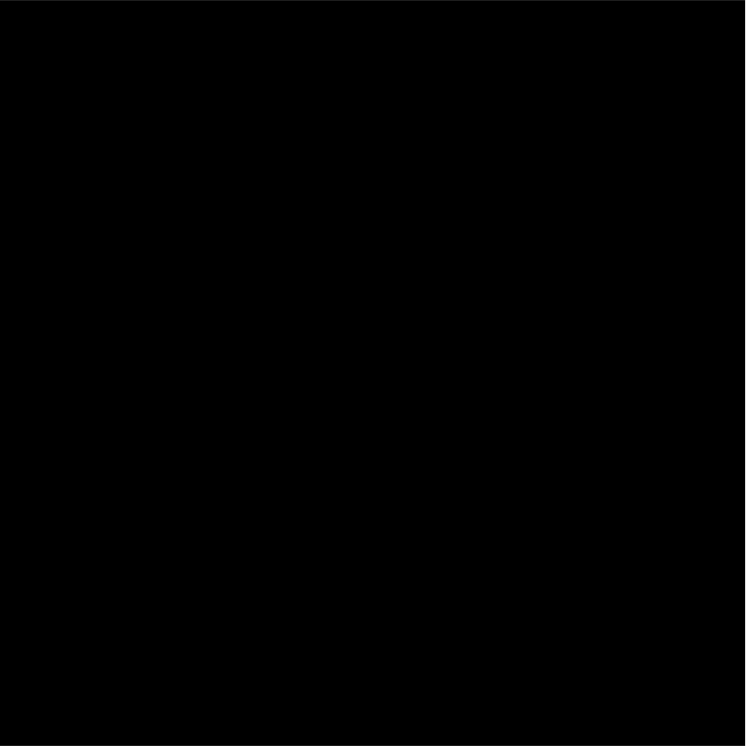} &
    \includegraphics[width=0.08\textwidth]{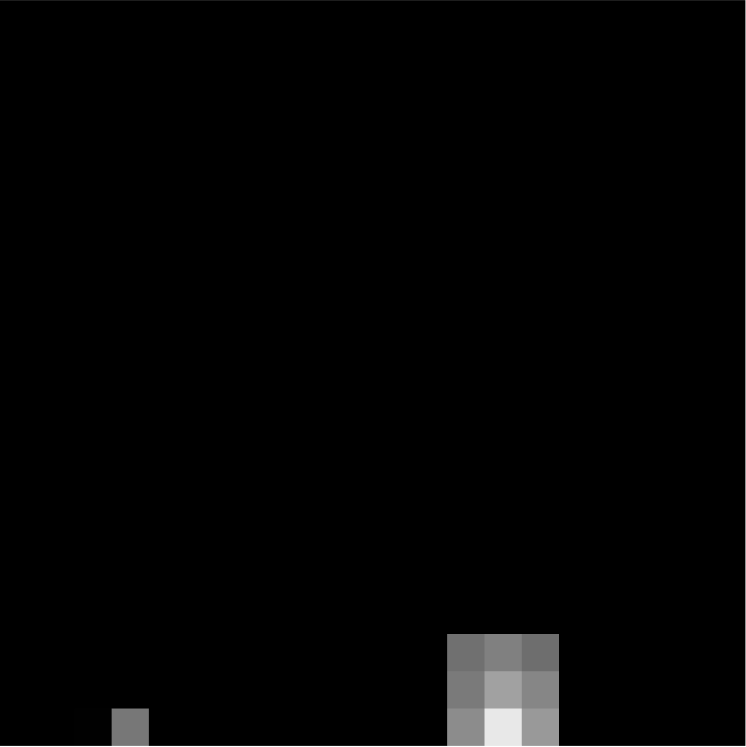} \\
  \end{tabular}  
  \caption{Semantic segmentation and BEV object detection for two scenes with the front view RGB image, ground truth semantic segmentation, semantic prediction, ground truth detection and detection prediction.}
  \label{fig:bev-decoder}
\end{figure}

Figure \ref{fig:bev-decoder} depicts the segmentation and BEV detection prediction beside the ground truth labels provided by CARLA from the decoder output, providing explainability. The segmentation can properly classify humans or traffic signs even in great distance and robust to environment perturbation such as shadows in different environments. However, it has limitations in capturing fine road details, such as crosswalk markings.

\begin{table*}[t]
    \vspace*{2mm} 
    \centering
    \renewcommand{\arraystretch}{1.2}
    \begin{adjustbox}{max width=\textwidth}
    \begin{tabular}{l|c|c|c|c|c|c|c|c|c}
        \hline
        \multirow{2}{*}{\textbf{Model}} & \multicolumn{3}{c|}{\textbf{LangAuto}} & \multicolumn{3}{c|}{\textbf{LangAuto-Short}} & \multicolumn{3}{c}{\textbf{LangAuto-Tiny}} \\
        & \textbf{DS ↑} & \textbf{RC ↑} & \textbf{IS ↑} & \textbf{DS ↑} & \textbf{RC ↑} & \textbf{IS ↑} & \textbf{DS ↑} & \textbf{RC ↑} & \textbf{IS ↑} \\
        \hline
        LMDrive Llama-7b \cite{shao2024lmdrive} & 31.3 &  37.1 & 0.82 & 42.8 & 49.1 & 0.87 & 52.2 & 57.8 & 0.91 \\
        LMDrive Llama (retrained) & 14.2 & 21.8 & 0.65 & 26.5  & 27.2 & \textbf{0.95} & 36.4 & 38.5 & \textbf{0.97} \\
        LMDrive LLaVA-7b \cite{shao2024lmdrive} & 36.2 & 46.5 & 0.81 & 50.6 & 60.0 & 0.84 & 66.5 & 77.9 & 0.85 \\
        \hline
        AD-H LLaVA-7b \cite{zhang2024ad} & 44.0 & 53.2 &\textbf{0.83} & 56.1 &  68.0 & 0.78 & \textbf{77.5} & \textbf{85.1} & 0.91 \\
        \hline
        Ours Llama-7b & \textbf{48.9} & \textbf{59.7} & 0.82 & \textbf{66.7} & \textbf{77.8} & 0.87 & 70.2 & 81.3 & 0.87 \\
        Ours Llama-7b (earlier checkpoint) & 39.6 & 50.0 & 0.81  & 63.4 & 68.0 & 0.94 & \textbf{77.5} & 80.0 & \textbf{0.97} \\
        Ours Llama3.1-8B-I& 33.19 & 40.77 & \textbf{0.83} & 60.96 & 65.86 & 0.92 & 66.0 & 69.98 & 0.90  \\
        \hline
        \end{tabular}
    \end{adjustbox}
    \caption{Comparison of BEVDriver against baseline models on the closed-loop LangAuto benchmark averaged on 3 repetitions per route.}
    \label{tab:baseline_comparison}
\end{table*}

\subsection{Quantitative Results}
\paragraph{Closed-Loop Planning}
Table \ref{tab:baseline_comparison} shows the results of our quantitative evaluation against the SoTA methods evaluated on the LangAuto benchmark introduced by LMDrive \cite{shao2024lmdrive}. BEVDriver demonstrates equal or improved performance compared to the best models in Driving Score (DS) and Route Completion (RC). In the LangAuto long benchmark, BEVDriver on Llama-7b surpasses the second best model, AD-H \cite{zhang2024ad}, by 11.1\% on the DS (48,9\%) and 12.2\% on RC (59.7\%). It also achieves best results on the LangAuto short by 18.9\% on the DS (66.7\%) and 14.5\% on RC (77.8\%), proving higher consistency and more stable driving. The Infraction Score (IS) matches state-of-the-art levels, despite the model staying on the route longer.
Compared to LMDrive's best model using LLaVA-7b, BEVDriver achieves a higher DS by 35.1\%, 31.2\% and 5.5\% on the long, short and tiny benchmarks, respectively.

We include LMDrive with Llama-7B backbone for direct architectural comparison. The table presents LMDrive’s published results \cite{shao2024lmdrive} alongside our retrained LMDrive version, evaluated on our frame rate and controller settings to ensure full comparability. BEVDriver surpasses the retrained LMDrive with same parameters on LangAuto on long, short and tiny by up to 244\% and leads by 56.2\% compared to their published results. 

An earlier training checkpoint of BEVDriver, trained for 10 epochs (48 hours), yields improved performance on LangAuto tiny regarding IS and DS, while being inferior on the short and long benchmark. This result indicates that further training improves the model towards fewer route deviations (better instruction following) and more consistent driving, while slightly increasing infractions, as reflected in a lower IS.

We further propose BEVDriver trained on the Llama3.1-8B-Instruct model, which is finetuned for instruction following, using the same training parameters of Llama-7b. Against expectations, the model performs worse on RC indicating lower performance on instruction following, but has higher IS on all benchmarks. 
Overall, the more powerful model does not perform better than Llama-7b. Whether this is due to suboptimal training parameters or other factors remains an open question for future investigation.

\paragraph{Open-Loop}
Our open-loop evaluation assesses the performance of trajectory prediction using ADE and FDE, comparing our model against LMDrive. BEVDriver Llama-7b achieves the best accuracy with an ADE of 0.04 and FDE of 0.07, followed by BEVDriver Llama-3.1-8B-Instruct model, which achieves an ADE of 0.08 and FDE of 0.13. In contrast, LMDrive Llama-7b and LMDrive LLaVA-7B yield ADE values of 0.09 and FDE values of 0.17 and 0.18, respectively, showing less precise trajectory predictions.

\subsection {Ablation Studies}
We perform several ablation studies.
\paragraph{Town 5 evaluation}
We train our model on a geofenced train/test split, leaving Town 5 out of the training data and run it on the official CARLA Town 5 long benchmark with the LangAuto instruction scheme. For comparison, Table \ref{tab:town05} includes LMDrive and our comparable models that have seen Town 5 during training. All models underperform on Town 5 compared to their LangAuto eight-town average. With a DS of 28.4\% and RC of 38.9\%, BEVDriver with Llama-7b performs best on this benchmark. BEVDriver’s higher Route Completion suggests better instruction adherence and driving stability. The LLama-3.1-8B-Instruct model has a slightly better IS but shorter RC than its Town 5-trained counterpart, indicating better generalizability on safe driving than instruction following.  
\begin{table}[ht]
    \centering
    \renewcommand{\arraystretch}{1.2}
    \begin{adjustbox}{max width=\columnwidth} 
    \begin{tabular}{l|c|c|c}
        \hline
        \multirow{2}{*}{\textbf{Model}} & \multicolumn{3}{c}{\textbf{Official Town05 Long}} \\
        \cline{2-4}
        & \textbf{DS ↑} & \textbf{RC ↑} & \textbf{IS ↑} \\
        \hline
        LMDrive Llama-7b (all)& 18.1& 20.2 & \textbf{0.94} \\
        Ours Llama-7b (all) & \textbf{28.4} & \textbf{38.9} & 0.73 \\
        Ours Llama-3.1 8B-I (all) & 25.59 & 30.71 & 0.82 \\
        Ours Llama-3.1 8B-I (no T5) & 22.84 & 27.64 & 0.83  \\
        \hline
    \end{tabular}
    \end{adjustbox}
    \caption{Evaluation on CARLA Town 5 long benchmark. Ablation on Llama-3.1-8B-Instruct trained without Town 5 (no T5).}
    \label{tab:town05}
\end{table}
\paragraph{End-to-End}
We conduct an ablation on full end-to-end training (Table \ref{tab:end-to-end}). After pre-training the encoder and pipeline in a modular way, we unfreeze the encoder to refine perception features, allowing it to train without traffic light status loss. Qualitative analysis confirms that the model learns to adhere to the traffic light status without this loss. The end-to-end training improves Route Completion by 5\%, suggesting enhanced navigation performance. The reduced infraction score might be due to pre-training on an encoder without traffic light status, worsening performance in the first place. We leave this interesting result open for future work as a promising avenue for full end-to-end training.
\begin{table}[ht]  
    \centering
    \renewcommand{\arraystretch}{1.2}
    \begin{adjustbox}{max width=\columnwidth}  
    \begin{tabular}{l|c|c|c}
        \hline
        \multirow{2}{*}{\textbf{Model}} & \multicolumn{3}{c}{\textbf{LangAuto-Tiny}} \\
        \cline{2-4}
        & \textbf{DS ↑} & \textbf{RC ↑} & \textbf{IS ↑} \\
        \hline
        Ours Llama-7b (frozen encoder) & \textbf{70.2} & 81.3 & \textbf{0.87} \\
        Ours Llama-7b (end-to-end) & 55.2 & \textbf{85.6} & 0.64 \\
        \hline
    \end{tabular}
    \end{adjustbox}
    \caption{Ablation of training BEVDriver end-to-end with unfrozen encoder.}
    \label{tab:end-to-end}
\end{table}

\paragraph{LangAuto without misleading instructions}
We analyze the impact of misleading instructions with results from LangAuto without misleading tasks shown in Table \ref{tab:no_misleading}. While these misleads should challenge the model, both LMDrive and our model perform worse without them. Further experiments indicate that more frequent navigation instructions improve the DS—66.8\% for BEVDriver and 46.05\% for LMDrive—compared to having none, suggesting a positive correlation between instruction frequency and Driving Score. 

\begin{table}[t] 
    \vspace*{2mm} 
    \centering
    \renewcommand{\arraystretch}{1.2}
    \begin{adjustbox}{max width=\columnwidth}
    \begin{tabular}{l|c|c|c}
        \hline
        \multirow{2}{*}{\textbf{Model}} & \multicolumn{3}{c}{\textbf{LangAuto-Tiny (no mislead)}} \\
        \cline{2-4}
        & \textbf{DS ↑} & \textbf{RC ↑} & \textbf{IS ↑} \\
        \hline
        LMDrive LLaVA 7B \cite{shao2024lmdrive} & 43.5& 47.9 & \textbf{0.90} \\
        Ours Llama-7b & \textbf{60.6} & \textbf{71.8} & 0.87 \\
        \hline
    \end{tabular}
    \end{adjustbox}
    \caption{Ablation without misleading instructions.}
    \label{tab:no_misleading}
\end{table}

\section{Discussion}
BEVDriver predicts low-level waypoints from latent BEV features and plans high-level maneuvers from natural language instructions, outperforming SoTA methods (AD-H \cite{zhang2024ad}) by up to 18.9\% on the LangAuto Driving Score. It surpasses LMDrive’s best model by up to 35.1\% and 56.2\% with the same backbone Llama-7b on the LangAuto benchmark and achieves generally better Driving Score and Route Completion on the long, short and tiny benchmark. We show that LLMs are capable of high waypoint prediction accuracy, while considering their inherent context understanding to incorporate natural language instructions using an efficient input representation through BEV maps. 
Our model performs well in closed-loop driving but struggles with distance-based instructions like "In x meters, turn left," executing them immediately at the next opportunity. Addressing this may require enhanced temporal reasoning, such as integrating spatiotemporal attention mechanisms or improving the alignment of distance-based navigation commands with the internal BEV representation.
Closed-loop evaluation is key to robust autonomous driving, with CARLA and LangAuto offering a solid LLM-based framework. However, limited scenario diversity and route-blocking traffic hinder generalization and metric reliability. 
Our proposed architecture extends beyond the autonomous driving domain and can be applied in various robotic application navigating complex environments, particularly those integrating natural language interfaces, like service and assistive robotics.
In future work, we aim to improve BEVDriver’s robustness and temporal awareness, e.g., through autoregressive prediction. Its flexible architecture enables extensions, such as generating language tokens for scene descriptions to enhance explainability. We have conducted first successful experiments on this extension. Additionally, end-to-end training warrants further evaluation. We plan to release the codebase as open source, enabling further research and development. 

\section{Conclusion}
We introduced BEVDriver, an end-to-end LLM-based motion planner that leverages a unified latent BEV representation to predict low-level waypoints and plan high-level maneuvers from natural language instructions. By fusing camera and LiDAR data, BEVDriver generates accurate future trajectories while incorporating context-driven decisions like turning and yielding. Our model outperforms state-of-the-art methods by up to 18.9\%, surpassing LMDrive’s best model by up to 35.1\% and achieving a 56.2\% improvement using the same Llama-7b backbone on the LangAuto benchmark. Future work will focus on enhancing BEVDriver’s robustness, temporal awareness, and explainability.

\section*{ACKNOWLEDGMENT}
The research leading to these results is funded by the German Federal Ministry for Economic Affairs and Climate Action within the project “NXT GEN AI METHODS – Generative Methoden für Perzeption, Prädiktion und Planung" (grant no. 19A23914M). The authors are solely responsible for the content of this publication.
\balance
\bibliographystyle{ieeetr}

\bibliography{references.bib}

\end{document}